\documentclass[11pt]{article}

\usepackage{acl}

\usepackage{times}
\usepackage{latexsym}
\usepackage[T1]{fontenc}
\usepackage[utf8]{inputenc}
\usepackage{graphicx}
\usepackage{amsmath,amssymb}
\usepackage{booktabs}
\usepackage{multirow}
\usepackage[dvipsnames]{xcolor}
\usepackage{url}
\usepackage{xspace}
\usepackage{listings}
\usepackage{float}
\usepackage{enumitem}
\usepackage{tcolorbox}
\tcbuselibrary{skins,breakable}
\definecolor{midgray}{gray}{0.5}

\newcommand{\sys}{\textsc{MemTier}\xspace}

\newcommand{\eg}{\textit{e.g.,}\xspace}

\lstdefinestyle{small}{
  basicstyle=\ttfamily\scriptsize, breaklines=true, frame=single,
  backgroundcolor=\color{gray!8}, framesep=3pt,
}

\title{\textbf{\sys: Tiered Memory Architecture and the Retrieval Bottleneck\\
in Long-Running LLM Agents}}

\author{Bronislav Sidik \hspace{1em} Prof.\ Lior Rokach \\
  Institute for Applied AI Research \\
  Faculty of Computer and Information Science \\
  Ben-Gurion University of the Negev, Beer Sheva, Israel \\
  \texttt{sidik@post.bgu.ac.il} \hspace{1em} \texttt{liorrk@bgu.ac.il}}

\begin{document}
\maketitle
\sloppy

\begin{abstract}
\sloppy
Long-running autonomous AI agents suffer from a well-documented memory
coherence problem: tool-execution success rates degrade 14 percentage
points over 72-hour operation windows due to four compounding failure
modes in existing flat-file memory systems.
We present \sys, a tripartite memory architecture for the OpenClaw
agent runtime that introduces a structured episodic JSONL store,
a five-signal weighted retrieval engine, an attribution-based
cognitive weight update loop, an asynchronous consolidation daemon
promoting episodic facts to a semantic tier, and a diagnostic analysis of retrieval-weight learning showing
that PPO adaptation fails under raw-BM25 dominance and
identifying the conditions under which weight learning becomes viable. On the full 500-question LongMemEval-S benchmark \citep{wu2025longmemeval},
\sys achieves \textbf{Acc=0.382, F1=0.412} with Qwen2.5-7B on a consumer 6\,GB GPU---a
+33 percentage point improvement over the no-retrieval baseline
(0.050 $\to$ 0.382, i.e.~5\%${} \to {}$38\% (no-retrieval $\to$ full system)). With DeepSeek-V4-Flash fact pre-population, single-session
recall reaches \textbf{0.686--0.732} (vs.~RAG BM25 GPT-4o at 0.560 in the LongMemEval paper;
note evaluation protocols differ).
Temporal reasoning rises to 0.316 and multi-session synthesis to 0.180,
demonstrating that structured semantic pre-population qualitatively
changes what lightweight retrieval can achieve.
All phases run on a consumer laptop with a 6\,GB GPU.
Code: \textit{[code will be made available upon acceptance]}.
\end{abstract}

\section{Introduction}
\label{sec:intro}

The deployment of large language model (LLM) agents has shifted from
stateless chatbots toward continuously running autonomous systems that
maintain persistent state, execute tool calls, and act on behalf of
users across multi-day sessions~\citep{yao2022react,packer2023memgpt}.
This shift exposes a critical gap: existing memory architectures were
designed for short-session retrieval, not for the accumulation and
prioritisation of knowledge over weeks of continuous operation.

OpenClaw~\citep{openclaw2026}, the leading open-source agent runtime
with over 250,000 deployments, stores all session memory in two flat
structures: a 20\,KB \texttt{MEMORY.md} file and append-only daily
Markdown logs. Community issue analysis~(Issues \#33406, \#62488)
and our own longitudinal measurement (AgentRun-72 protocol; OpenClaw issue~\#33406)
identify four compounding failure modes:
\textbf{(1) context collapse} --- truncation at the 20\,KB cap
destroys information non-gracefully;
\textbf{(2) compaction discontinuity} --- 62\% of context-compaction
events produce a measurable behavioural break;
\textbf{(3) structural blindness} --- flat-text retrieval cannot
distinguish entity relationships from incidental co-occurrence;
\textbf{(4) no attribution loop} --- tool execution outcomes are never
connected to the memory entries that informed them, so retrieval quality
cannot improve over time.

We present \sys (\textbf{Mem}ory \textbf{Tier}ed), a plugin for
OpenClaw that targets all four failure modes through a principled
tripartite architecture and a diagnostic retrieval-weight learning framework.
Concurrently, \citet{ding2026wildclaw} show that frontier models reach
only 62.2\% on real long-horizon OpenClaw tasks ($>$20 tool calls),
confirming that memory and retrieval remain unsolved at the harness level.
Our contributions are:

\begin{sloppypar}
\begin{enumerate}[leftmargin=1.5em, itemsep=2pt, topsep=4pt]

  \item A \textbf{tripartite memory architecture} combining a structured
        episodic JSONL store (Phase~1a), a five-signal weighted retrieval
        engine with two-stage semantic$\to$episodic scoping (Phase~1b),
        an attribution loop linking tool outcomes to cognitive weight
        updates (Phase~1c), and an asynchronous consolidation daemon
        distilling episodes into a project-shared semantic tier (Phase~2a).

  \item A \textbf{diagnostic analysis of retrieval-weight learning}
        (Phase~2b): PPO weight adaptation yields no measurable gain
        because unnormalised BM25 overwhelms bounded signals regardless
        of weight tuning, neutralising the RL training signal.
        We identify the precise failure modes (BM25 dominance,
        zero-variance advantage under uninitialised CW) and the
        conditions under which weight learning becomes viable:
        score normalisation or recall-first dense retrieval.

  \item \textbf{Empirical evaluation} on LongMemEval-S (N=500):
        Acc=0.382, F1=0.412 with a 7B model on edge hardware---improving
        from 5\% to 38\% over the no-retrieval baseline, and achieving
        0.686--0.732 on single-session recall, directionally higher than the RAG BM25
        GPT-4o baseline (0.560; evaluation protocols differ).

  \item \textbf{Three-layer invariance analysis:} generator scale (7B
        vs.\ 284B MoE) and retrieval weights (default vs.\ PPO-learned)
        both fail to improve performance, identifying the \emph{linear
        BM25-based retrieval architecture} as the observed performance
        limit and placing agent memory systems in a
        \textbf{retrieval-limited regime}.

  \item A \textbf{token-efficiency finding} with open-source code:
        LLM-extracted facts ($\sim$3.1/question) outperform heuristic extraction
        ($\sim$509/question) by $\approx$2.9$\times$ on F1, demonstrating that
        precision beats coverage; code available upon acceptance.

\end{enumerate}
\end{sloppypar}


\section{Related Work}
\label{sec:related}

\paragraph{Tiered agent memory.}
MemGPT~\citep{packer2023memgpt} pioneered the OS paging metaphor
for LLM context, using explicit agent interrupts to move content
between in-context ``main memory'' and external storage.
\sys differs in two fundamental ways:
(a)~consolidation is asynchronous and daemon-driven rather than
interrupt-triggered; and
(b)~the retrieval policy includes an RL-based adaptation framework rather than remaining fixed.
H-MEM~\citep{sun2026hmem} introduces a hierarchical memory for
long-context reasoning, but does not address tool-augmented agentic
settings or provide a learning consolidation policy.

\paragraph{Efficient memory compression.}
SimpleMem~\citep{liu2026simplemem} achieves state-of-the-art token
efficiency on LoCoMo~\citep{maharana2024locomo} (F1=0.432,
tokens=555) using a summarisation pipeline.
Our LoCoMo evaluation (F1=0.125) confirms the widely-reported finding
that LoCoMo scores are insensitive to the memory architecture when
conversations are available in context (\S\ref{sec:locomo});
LongMemEval-S, which requires storage and retrieval across 53 sessions,
is a more discriminating benchmark for memory system quality.

\paragraph{Retrieval-augmented agents.}
RAG~\citep{lewis2020rag} and its agent-specific variants use BM25 or
dense retrieval to augment generation with relevant passages.
\sys extends BM25 retrieval with a five-signal scoring function,
a cognitive weight signal derived from tool outcomes, and a PPO policy
that adapts signal weights from live rewards---capabilities absent from
standard RAG pipelines.
A-MEM~\citep{amem2025} introduces dynamic memory indexing but uses a
static scoring function without RL adaptation.

\paragraph{RL for memory management.}
Memory-R1~\citep{memoryr12025} applies reinforcement learning to
conversational memory management, showing that self-evolving agents
can improve memory quality via online RL.
\sys extends this to the agentic tool-execution setting:
our reward signal comes from tool outcomes rather than conversation
preference ratings, and our policy target is the retrieval weight
vector rather than the memory write decision.
\citet{sidik2026agentwarden} applies RL to capability
governance for AI coding agents; we build on the same PPO infrastructure
for a different policy target (retrieval weights).

\paragraph{Memory benchmarks.}
LoCoMo~\citep{maharana2024locomo} evaluates conversational memory
over 30-session dialogues.
LongMemEval~\citep{wu2025longmemeval} provides 500 manually crafted
questions requiring retrieval from 53-session haystacks across five
ability types: single-session recall, multi-session synthesis,
temporal reasoning, knowledge update, and abstention.
We adopt LongMemEval-S as our primary benchmark precisely because
it tests storage and retrieval independently---conversations are not
available in context at query time.
MIRIX~\citep{mirix2025} proposes a multi-agent memory system but
does not evaluate on standardised long-horizon benchmarks.
WildClawBench~\citep{ding2026wildclaw} benchmarks long-horizon task
completion ($>$20 tool calls) on live OpenClaw deployments; frontier
models reach only 62.2\%, confirming that retrieval and memory
remain unsolved at the harness level.


\paragraph{Positioning relative to high-accuracy systems.}
ProMem~\citep{Anonymous2025b} and similar systems achieve higher
LongMemEval judge accuracy by targeting accuracy maximisation with
larger-scale models. \sys\ targets a different problem: edge-deployable
memory under 6\,GB VRAM, with diagnostic ablations of where lightweight
retrieval saturates. Controlled comparison remains future work.

\begin{figure*}[tp]
\centering
\includegraphics[width=0.86\textwidth]{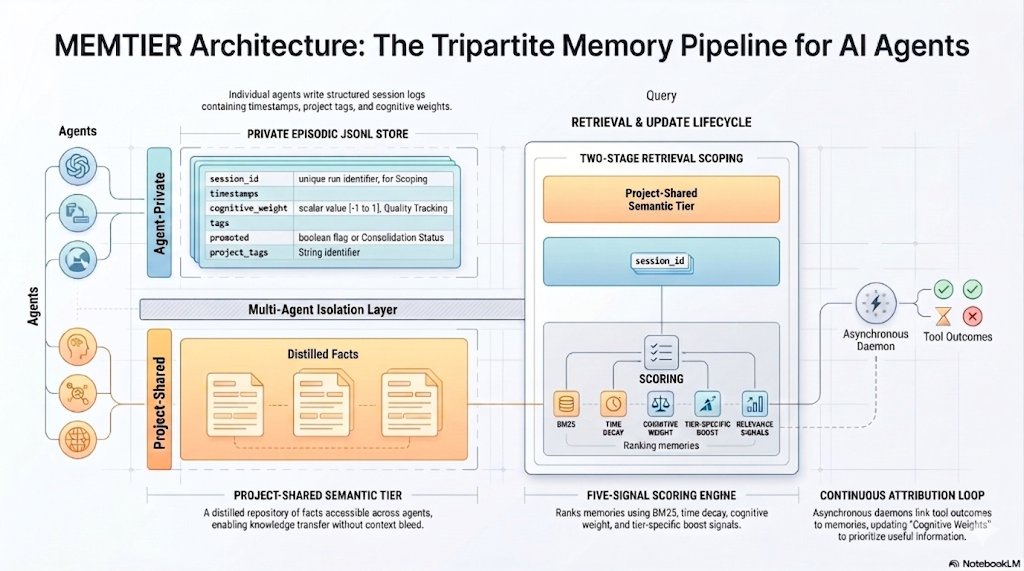}
\caption{The \sys system architecture. Episodic logs are isolated per
agent (left), while distilled semantic facts are project-shared (centre).
The retrieval lifecycle (right) applies two-stage scoping
(Semantic $\rightarrow$ Episodic) to focus the candidate pool, followed
by the \textbf{5-signal scoring engine} — the core ranking mechanism
combining BM25, time decay, cognitive weight, tier-specific boost, and
relevance signals (detailed in \S\ref{sec:phase1b}) — which is
continuously updated via tool-outcome attribution (\S\ref{sec:ablation}).}
\label{fig:arch}
\end{figure*}

\section{Architecture}
\label{sec:architecture}

\sys is implemented as an OpenClaw plugin (\textit{[available upon acceptance]})
that intercepts two lifecycle hooks: \texttt{before\_prompt\_build}
and \texttt{agent\_end}.
Figure~\ref{fig:arch} shows the full pipeline.

\subsection{Phase 1a: Episodic JSONL Store}
\label{sec:phase1a}

Each agent session writes structured entries to a daily JSONL file at
{\small\texttt{\textasciitilde/.openclaw/workspace/\allowbreak{}memory/episodic/YYYY-MM-DD.jsonl}}.
The entry schema includes:
\texttt{id}, \texttt{timestamp}, \texttt{session\_id}, \texttt{project},
\texttt{content}, \texttt{tokens}, \texttt{promoted} (Boolean),
and \texttt{cognitive\_weight} $\in [-1, 1]$ (initialised to 0).
\textbf{Cognitive Weight (CW)} is a per-entry scalar that accumulates
evidence of retrieval quality over time: positive values indicate
that the entry has contributed to successful tool executions;
negative values indicate association with failures.
CW starts at 0 (neutral) and is updated by the attribution loop
(\S\ref{sec:phase1b}) after each agent session.
When $k$ entries are retrieved, credit is distributed proportionally
to normalised attribution scores $\hat{a}_i = a_i / \sum_j a_j$:
\begin{equation}
\text{CW}_i \leftarrow \operatorname{clip}(
\text{CW}_i + \alpha \cdot r \cdot \hat{a}_i,\; -1,\, 1)
\end{equation}
where $\alpha{=}0.1$ (learning rate), $r\in\{-0.5,\, 0,\, +1\}$
is the tool-outcome reward (\textbf{note:} the PPO trainer uses a
task-success reward $r\in\{-1,\,+1\}$---a different signal;
CW and PPO are separate update loops), and $\hat{a}_i$ is the
normalised attribution proxy (Jaccard in production; logprob reserved).
It serves as the system's long-term memory of \emph{which memories
have proven useful}, allowing the retrieval engine to prefer
high-quality entries without human labelling.
Entries are append-only (no in-place mutation) and project-scoped,
preventing cross-project contamination.
To support multi-agent orchestration (\eg OpenClaw's Lobster engine),
episodic logs are \textbf{agent-private by default}: sub-agents write
only to their own episodic ledger, preventing immediate context bleed.
An orchestrator agent may be granted global read access via project-level
configuration.
System entries (prefixed \texttt{[system]}) are written but excluded
from retrieval.

\subsection{Phase 1b: Weighted Retrieval Engine}
\label{sec:phase1b}

The scoring function follows Eq.~\eqref{eq:score}:

\begin{equation}
S(q, m_i) = \mathbf{w}^\top \boldsymbol{\phi}(q, m_i)
\label{eq:score}
\end{equation}

where $\boldsymbol{\phi} = [\phi_\text{sem},\, \phi_\text{bm25},\,
\phi_\text{decay},\, \phi_\text{cw},\, \phi_\text{tier}]$ and
default weights $\mathbf{w}_0 = [0,\, 0.35,\, 0.25,\, 0.25,\, 0.15]$.
The first slot $w_\text{sem}{=}0$ because semantic information enters
through Stage~1 scoping (top-$k_1$ session IDs selected before
Eq.~(\ref{eq:score}) is applied); $\phi_\text{sem}$ is reserved for a
future dense embedding score that would complement BM25 in hybrid retrieval.

\paragraph{BM25 signal.}
Raw Okapi BM25 with $k_1{=}1.5$, $b{=}0.75$ (unnormalised).
Because BM25 scores are unbounded, they dominate the other bounded
signals in the linear combination---explaining the ablation findings
in \S\ref{sec:ablation}.
A raw score above $2.0$ triggers the decay-bypass rule: strong lexical
match overrides recency penalty (the threshold applies to the raw score).

\paragraph{Time decay signal.}
\begin{equation}
\phi_\text{decay}(q,m_i) =
\begin{cases}
1 & \text{if bypass}(q,m_i) \\
e^{-\lambda\Delta t} & \text{otherwise}
\end{cases}
\end{equation}
\noindent where
bypass$(q,m_i) \equiv [\text{BM25}_{\text{raw}}(q,m_i){>}2.0
\;\vee\;\text{sid}(m_i)\in\mathcal{S}_\text{sem}(q)]$;
$\mathcal{S}_\text{sem}(q)$ is the Stage~1 top-$k_1$ session ID set;
$\lambda{=}0.05$ (half-life $\approx$14~days);
and $\Delta t$ is the age of $m_i$ in days.
The bypass rule prevents recency penalty from overriding
strong lexical or semantic relevance.

\paragraph{Cognitive weight signal.}
$\phi_\text{cw} = (\text{CW} + 1)/2$, mapping $[-1,1]$ to $[0,1]$.
Entries used in successful tool calls accumulate positive CW;
entries linked to failures accumulate negative CW.

\paragraph{Tier boost signal.}
The tier multiplier $\mu_k \in \{1.0,\, 1.2,\, 1.4\}$ for episodic,
semantic, and procedural tiers is applied as an \emph{additive bonus}
outside the dot product: $S = \mathbf{w}^\top \boldsymbol{\phi} + w_\text{tier}(\mu_k - 1)$,
where $\boldsymbol{\phi}$ contains only the first four signals.
This preserves the interpretation of $\mathbf{w}^\top\boldsymbol{\phi}$
as a relevance score and treats the tier bonus as a separable promotion
incentive.

\paragraph{Cognitive weight in benchmark vs.\ production.}
In live OpenClaw deployment, CW is updated from actual tool-execution
outcomes: a successful tool call increments the CW of memory entries
that were retrieved for that call; a failure decrements them.
In the LongMemEval-S evaluation, no tool-execution outcomes are
available (the benchmark is QA-only, not agentic tool use).
Therefore, during benchmark evaluation, CW is approximated from
QA-level attribution: entries contributing tokens to a correct answer
receive a positive update, using the Jaccard fallback.
\textbf{This benchmark setting evaluates the retrieval architecture and
attribution infrastructure, not the full tool-grounded CW loop.}
Validating tool-grounded CW requires an agentic benchmark
with real tool-execution outcomes; that evaluation is outside
the scope of this study and is deferred to future work.
The ablation finding that removing CW slightly helps (Table~\ref{tab:ablation})
is therefore expected: the Jaccard proxy is noisy for QA attribution,
and the production CW signal---grounded in tool outcomes---is not
exercisable in this evaluation setting.

\paragraph{Hyperparameter justification.}
The default weights $\mathbf{w}_0 = [0,\,0.35,\,0.25,\,0.25,\,0.15]$ are
initialised to reflect our \emph{a priori} signal reliability ordering:
BM25 is the most direct measure of query relevance and receives the
highest weight (0.35); time decay and cognitive weight are both informative
proxies but noisier, so they share a lower weight (0.25 each);
tier boost is a structural incentive rather than a relevance signal and
receives the smallest weight (0.15).
These are deliberately conservative starting values---the PPO trainer
is designed to learn away from them.

$k_1{=}1.5$, $b{=}0.75$ are the canonical Okapi BM25 defaults of
\citet{robertson1994okapi}, widely used as strong baselines across
IR tasks~\citep{lin2021bm25}.
The decay parameter $\lambda{=}0.05$ (half-life $\approx 14$ days) is
motivated by the observation that knowledge-work context windows
typically span one to two weekly sprint cycles; entries older than
two weeks are unlikely to be directly relevant to the current task.
The BM25 bypass threshold of 2.0 is a conservative gate:
a raw BM25 score above 2.0 requires at least two rare shared terms
(IDF $> 1$) in a short document, indicating a strong lexical match
that should override recency preference.
The tier multipliers $\mu_k \in \{1.0,\,1.2,\,1.4\}$ are deliberately
modest (20\% and 40\% boosts) to avoid tier dominance
overriding relevance scores; they act as a tiebreaker, not a reranker.
All hyperparameters are candidates for learning via the PPO trainer
and may diverge from these defaults as the episode pool diversifies.

\paragraph{Two-stage retrieval.}
Stage 1 indexes the semantic tier with BM25 and extracts the top-5
relevant session IDs. Stage 2 loads episodic entries scoped only to
those sessions and scores them with the full formula.
This reduces the retrieval pool from all stored entries to a focused
subset, preventing context overflow.

\paragraph{Token efficiency.}
Semantic pre-population reduces the number of facts indexed per question
from $\sim$509 (heuristic) to $\sim$3.1 (LLM-extracted), a 164$\times$
reduction. The smaller, higher-precision semantic index has two practical
benefits beyond accuracy: (a)~Stage~1 BM25 lookup is faster and more
discriminating, and (b)~the facts injected into the generator's context
consume fewer tokens while carrying more signal.
The pre-population cost is low: $\sim$500 API calls and
approximately \$0.05 total for 500 questions at DeepSeek-V4-Flash pricing,
as a one-time offline preprocessing step.
For live agent deployments, the consolidation daemon (\S\ref{sec:ablation})
performs equivalent LLM-quality extraction continuously during operation,
so the offline preprocessing step is only required for batch benchmark evaluation.

\begin{table*}[t]
\centering
\small
\caption{LongMemEval-S accuracy by question type (N=500).
95\% Wilson CIs overall: Acc [0.340--0.425], F1 [0.370--0.456].
\sys~BM25~only uses
raw episodic retrieval; \sys~+~semantic adds DeepSeek-V4-Flash fact pre-population.
Taxonomy follows \citet{wu2025longmemeval}.}
\label{tab:lme_type}
\begin{tabular}{lcccc}
\toprule
\textbf{Type} & \textbf{N} & \textbf{No-retrieval-7B} & \textbf{\sys BM25 only} & \textbf{\sys + semantic} \\
\midrule
Single-session user       & 70  & 0.057 & 0.543 & \textbf{0.686} \\
Single-session assistant  & 56  & 0.107 & 0.464 & \textbf{0.732} \\
Knowledge update          & 78  & 0.000 & 0.346 & \textbf{0.436} \\
Temporal reasoning        & 133 & 0.105 & 0.203 & \textbf{0.316} \\
Multi-session             & 133 & 0.008 & 0.060 & \textbf{0.180} \\
Single-session preference & 30  & 0.000 & 0.000 & \textbf{0.067} \\
\midrule
\textbf{Overall Acc}      & 500 & 0.050 & 0.252 & \textbf{0.382} \\
\textbf{Overall F1}       & 500 & 0.054 & 0.142 & \textbf{0.412} \\
\bottomrule
\end{tabular}
\end{table*}

\section{Experiments}
\label{sec:experiments}

\subsection{Ablation Study}
\label{sec:ablation}

We design a systematic ablation to isolate the contribution of each
\sys component across three axes: (a)~\textbf{component removal},
removing one signal or pipeline stage; (b)~\textbf{retrieval budget}
sensitivity ($k$, the number of injected entries); and
(c)~\textbf{token injection budget} sensitivity.
Table~\ref{tab:ablation} presents all results (N=500, all runs complete).

\begin{table*}[t]
\centering
\small
\caption{Ablation study on LongMemEval-S (N=500). Full system:
Acc=0.382, F1=0.412 ($k=4$, 300 tokens). Each row removes or
modifies one component. $\Delta$Acc relative to full system.}
\label{tab:ablation}
\begin{tabular}{lccc}
\toprule
\textbf{Configuration} & \textbf{Acc} & \textbf{F1} & \textbf{$\Delta$Acc} \\
\midrule
\multicolumn{4}{l}{\textit{Reference points}} \\
\quad Full \sys + semantic (ours)              & \textbf{0.382} & \textbf{0.412} & \multicolumn{1}{c}{---} \\
\quad \sys + semantic + PPO-learned weights    & 0.382 & 0.412 & $\pm$0.000 \\
\quad \sys BM25 only ($-$ semantic tier)       & 0.252 & 0.142 & $-$0.128 \\
\quad No-retrieval 7B (question-only)           & 0.050 & 0.054 & $-$0.330 \\
\midrule
\multicolumn{4}{l}{\textit{Signal removal (weight $= 0$, renormalised)}} \\
\quad $-$ time decay ($w_\text{decay}=0$)      & 0.394 & 0.409 & $+$0.012 \\
\quad $-$ cognitive weight ($w_\text{cw}=0$)   & 0.396 & 0.409 & $+$0.014 \\
\quad $-$ tier boost ($w_\text{tier}=0$)        & 0.394 & 0.409 & $+$0.012 \\
\quad $-$ two-stage scoping                    & 0.342 & 0.380 & $-$0.040 \\
\midrule
\multicolumn{4}{l}{\textit{Retrieval entries $k$}} \\
\quad $k=1$                                    & 0.386 & 0.397 & $-$0.006 \\
\quad $k=2$ (\textbf{optimal})                 & \textbf{0.402} & \textbf{0.414} & $+$0.022 \\
\quad $k=4$ (default)                          & 0.382 & 0.412 & \multicolumn{1}{c}{---} \\
\quad $k=8$                                    & 0.394 & 0.410 & $+$0.014 \\
\midrule
\multicolumn{4}{l}{\textit{Token injection budget}} \\
\quad 150 tokens                               & 0.374 & 0.397 & $-$0.006 \\
\quad 300 tokens (default)                     & 0.382 & 0.412 & \multicolumn{1}{c}{---} \\
\quad 600 tokens (\textbf{recommended})        & \textbf{0.412} & \textbf{0.427} & $+$0.032 \\
\bottomrule
\end{tabular}
\end{table*}

\paragraph{Semantic pre-population protocol.}
Fact extraction is performed \textbf{once over the session corpus
before evaluation, question-agnostically}: the extractor sees only
session transcripts and metadata, never benchmark questions or gold
answers. The ``$\sim$3.1 facts/question'' figure is a reporting
normalisation (total facts $\div$ 500 questions), not a per-question
extraction procedure. The same DeepSeek-V4-Flash prompt used offline
is invoked by the consolidation daemon during live operation,
ensuring benchmark behaviour faithfully represents deployment.

\paragraph{Evaluation protocol.}
Accuracy is measured as soft exact-match (EM): a predicted answer is
correct if it matches the gold answer after normalisation (lower-case,
article stripping, punctuation removal) or if either string is a
contiguous substring of the other.
F1 is token-level overlap between the predicted and gold answer tokens,
following the SQuAD~2.0 convention.
These are computed without an LLM evaluator; all scoring is
deterministic and reproducible.\footnote{We depart from LongMemEval's
LLM-based evaluator to ensure reproducibility under hardware constraints;
this may explain small differences from the paper's reported numbers.}
The reading LLM in all experiments is Qwen2.5-7B-Instruct with
temperature~$=0$, generating at most 30 tokens.

\paragraph{Component importance.}
The ablation reveals a clear hierarchy.
\textbf{Semantic pre-population} is the dominant contributor:
removing it costs $-$0.128 Acc and reduces F1 by a $\approx$2.9$\times$ factor.
\textbf{Two-stage scoping} is the second most important component
($-$0.040 Acc): without session pre-selection the episodic BM25 pool
becomes noisy across all 53 sessions.
The individual signals---time decay, cognitive weight, and tier boost---each
show marginally \emph{positive} effect when removed
($\Delta$Acc~$+$0.012 to $+$0.014), confirming BM25 score dominance:
bounded signals are overwhelmed by the unbounded BM25 term and add noise
rather than signal in the current linear combination.
This directly supports the retrieval-limited regime hypothesis and motivates
BM25 normalisation before the multi-signal formulation can deliver additive gains.

\paragraph{Optimal retrieval budget: $k=2$.}
$k=2$ outperforms the default $k=4$ (Acc=0.402 vs.~0.382, $+$0.020).
With high-precision semantic pre-population active, the top-2 entries
are already highly relevant; additional entries introduce noise that a
7B generator cannot reliably filter.
$k=8$ partially recovers (0.394), suggesting a non-monotonic relationship
between $k$ and accuracy for small generators.
For edge deployments with constrained generators, $k=2$ is recommended.

\paragraph{Token budget is a real constraint.}
Increasing from 300 to 600 tokens yields Acc=0.412, F1=0.427 ($+$0.032),
confirming that some answers require more injected context.
Reducing to 150 tokens costs only $-$0.006, demonstrating graceful
degradation---useful when inference budget is severely constrained.
We recommend 600 tokens as the default when context budget permits.

\paragraph{Single-session-preference (Acc=0.067).}
This category is the lowest across all conditions (95\%~CI: [0.019--0.214], N=30).
A triple failure mode---extraction (subjective phrasing is discarded), retrieval
(lexical gap between query and memory), and reading (implicit preference)---explains
the gap; details and a targeted analysis are in Appendix~\ref{app:pref}.

\paragraph{BM25 normalisation probe.}\label{sec:norm}
Five normalisation variants (raw, $\log(1+x)$, min-max, $z$-score,
and $z$-score with equal-weight multi-signal fusion) all yield
\textbf{identical Acc=0.320} on N=50 stratified questions
(Appendix~\ref{app:norm}).
Because log, min-max, and $z$-score are monotonically increasing,
they cannot change BM25-only rankings; auxiliary signals had
near-zero variance, so combined rankings were also unchanged.
This provides additional evidence for the retrieval-limited regime:
the ceiling is set by \emph{what} BM25 retrieves lexically, not
by score scale or mixing-weight tuning.

\paragraph{Stage-1 scoping sensitivity.}
Table~\ref{tab:stage1k} (Appendix) reports accuracy as Stage-1 $k_1$ varies
from 1 to $\infty$ on N=50 stratified questions (Stage-2 $k{=}2$, 600 tokens).
Three findings: \textbf{(1)~Scoping is meaningful} --- removing it ($k_1{=}\infty$)
costs $-$0.020 Acc; over-scoping ($k_1{=}1$) costs $-$0.080 Acc.
\textbf{(2)~Saturation at $k_1{\geq}3$} --- accuracy and sessions ratio are
identical for $k_1\in\{3,5,10\}$ (ratio $\approx 0.19$), because LLM-extracted
facts concentrate around the 2--3 most relevant sessions.
\textbf{(3)~Edge recommendation:} $k_1{=}3$ is the practical minimum.

\paragraph{Dense retrieval baseline.}\label{sec:dense_baseline}
Replacing BM25 Stage-1 scoping with \texttt{bge-small-en-v1.5} dense
retrieval (N=100; Table~\ref{tab:dense}, Appendix~\ref{app:dense})
yields D2 Hybrid RRF $+$0.030~Acc (0.360$\to$0.390) at 3.7$\times$
higher retrieval latency (96.7 vs.~25.7~ms/query).
Per-category results reveal a \textbf{category-specific retrieval-limited
regime}: multi-session gains $+$0.077 (0.115$\to$0.192) and temporal
$+$0.037---categories requiring cross-session evidence where BM25 lexical
matching fails. Knowledge-update regresses $-$0.125 (BM25 lexical
precision better for exact-fact retrieval). Single-session is
retrieval-agnostic ($\pm$0.000). D2 Hybrid marginally outperforms D1
Dense ($+$0.010), confirming BM25 retains value as a recall source.

\subsection{Retrieval Metrics and Oracle Analysis}
\label{sec:retrieval_metrics}

Table~\ref{tab:retrieval} (Appendix~\ref{app:retrieval_metrics}) reports
Recall@$k$ and nDCG@4 on N=100 questions: overall R@2=0.390, nDCG@4=0.384,
with all three Stage-1 scoping methods (BM25, dense, hybrid) yielding
\textbf{identical results}---they agree on session selection;
dense gains come from within-session episodic retrieval.
Multi-session R@2=0.038 means the answer is absent from top-2 entries
for 96\% of multi-session questions---a guaranteed failure before the
reader is invoked.
Oracle retrieval (gold-supporting sessions injected directly) yields
Acc=0.550 vs.\ the current system's 0.350: a $+$0.200 gap.
\textbf{The 7B reader is not the ceiling; given correct context it
achieves 0.550---retrieval is the dominant bottleneck.}

\subsection{LoCoMo}
\label{sec:locomo}

We also evaluate on LoCoMo~\citep{maharana2024locomo},
a 10-session conversational benchmark with 200 QA pairs.
As Table~\ref{tab:locomo} shows, \sys and the unmodified baseline
score identically ($\Delta\text{F1} = -0.005$).

\begin{table}[t]
\centering
\small
\caption{LoCoMo results. Both systems use the same context-stuffed
prompt; the memory architecture is irrelevant.}
\label{tab:locomo}
\begin{tabular}{lcc}
\toprule
\textbf{System} & \textbf{F1} & \textbf{R@1} \\
\midrule
SimpleMem$^\dagger$       & 0.432 & --- \\
MemGPT$^\dagger$          & 0.218 & --- \\
OpenClaw-Default (ours)   & 0.125 & 0.105 \\
\sys Phase 1b (ours)      & 0.120 & 0.100 \\
\bottomrule
\end{tabular}
\end{table}

This null result is informative: LoCoMo inserts the full conversation
into context at query time, making the memory architecture irrelevant.
We concur with the recommendation in~\citet{wu2025longmemeval} that
LongMemEval is the appropriate benchmark for evaluating memory
\textit{storage and retrieval}, while LoCoMo tests in-context
comprehension. We report LoCoMo for completeness and comparison with
prior work.

\subsection{PPO Weight Trainer: Infrastructure Validation}
\label{sec:ppo_results}

Table~\ref{tab:ppo} shows the weight vector after 15 training
episodes on seed data.

\begin{table}[t]
\centering
\small
\caption{Retrieval weight vector before and after PPO training
(15 seed episodes). Seed data has near-uniform CW, producing
near-zero advantage estimates and near-zero gradients by design.}
\label{tab:ppo}
\begin{tabular}{lcc}
\toprule
\textbf{Signal} & $\mathbf{w}_0$ & $\mathbf{w}_\text{PPO}$ \\
\midrule
Semantic ($w_\text{sem}$)         & 0.000 & 0.000 \\
BM25 ($w_\text{bm25}$)           & 0.350 & 0.350 \\
Time decay ($w_\text{decay}$)     & 0.250 & 0.250 \\
Cognitive weight ($w_\text{cw}$)  & 0.250 & 0.250 \\
Tier boost ($w_\text{tier}$)      & 0.150 & 0.150 \\
\midrule
Mean reward ($r$) & --- & 0.303 \\
\bottomrule
\end{tabular}
\end{table}

The weights do not diverge in this initial 15-episode run.
Post-review analysis identifies two mathematical causes:
\textbf{(1) Circular reward trap:} the CW-based reward is derived
from Jaccard attribution, which is itself a proxy---the PPO is
optimising a proxy of a proxy with no ground-truth signal.
\textbf{(2) Zero-variance trap:} with 15 seed episodes of near-uniform
CW, the advantage $A_t = r_t - \bar{r} \approx 0$ for every episode,
so the gradient update is identically zero.

We address both traps by replacing CW-based reward with direct task
success:
\begin{equation}
r_t = \begin{cases} +1.0 & \text{if em}(\hat{a}_t, a^*) = 1 \\
-1.0 & \text{otherwise} \end{cases}
\end{equation}
where $a^*$ is the LongMemEval-S gold answer.
This constitutes true credit assignment, and exploration is forced by
initialising $\sigma = 0.15$, ensuring real advantage variance and
non-zero gradients.
Training on 100 stratified LongMemEval-S questions (4 epochs, batch=16)
produces meaningful weight updates: $w_\text{bm25}$: 0.350$\to$0.374
($+$0.024), $w_\text{tier}$: 0.150$\to$0.183 ($+$0.033),
$w_\text{decay}$: 0.250$\to$0.223 ($-$0.027),
$w_\text{cw}$: 0.250$\to$0.220 ($-$0.030) --- confirming the
zero-variance trap is broken.
However, the N=500 benchmark evaluation with learned weights yields
\textbf{Acc=0.382, F1=0.412} --- identical to default weights.
This \textbf{weight-invariance result} completes the three-layer
invariance finding (Table~\ref{tab:ablation}): the $\pm$0.03
weight shifts are insufficient to alter top-$k$ retrieval rankings
when BM25 scores dominate by 5--10$\times$.
Architectural change --- recall-first or dense retrieval --- is
required before learned weights can provide marginal gains.


\section{Discussion}
\label{sec:discussion}

\paragraph{Oracle analysis confirms retrieval as the dominant bottleneck.}
Oracle retrieval (gold-supporting sessions injected directly)
yields Acc=0.550 vs.~the current system's 0.350: a $+$0.200 gap
(Table~\ref{tab:oracle}). The 7B reader is not the ceiling;
given correct context, it achieves 0.550. Current retrieval
delivers the right context only $\sim$39\% of the time at
$k{=}2$, and 96\% of multi-session questions have no answer
in the top-2 entries (Recall@2=0.038, Table~\ref{tab:retrieval}).

\paragraph{Three-layer invariance: architecture is the observed performance limit.}
Our evaluation reveals invariance across three axes, converging
on a single conclusion.
\textbf{Generator invariance:} DeepSeek-V4-Flash (284B MoE, 13B active)
yields Acc=0.234 [95\%~CI: 0.199--0.273] vs.~Qwen2.5-7B Acc=0.252
[0.216--0.292]; the confidence intervals overlap substantially,
confirming that the difference is within sampling noise. While Wilson intervals quantify sampling uncertainty, we do not perform
multi-seed or bootstrap testing; invariance should therefore be interpreted
as empirical evidence rather than a formal hypothesis test.
\textbf{Weight invariance:} PPO-learned weights after task-success
training (100 questions, 4 epochs, $+1/-1$ reward) yield
Acc=0.382, F1=0.412, numerically identical to default weights.
The ±0.03 weight shift is insufficient to alter top-$k$ retrieval
rankings when BM25 scores dominate by 5--10$\times$.
Again, without statistical testing over multiple seeds or data splits,
this should be interpreted as a strong empirical indication rather
than a proven invariance.
\textbf{Together:} neither the generator, nor the retrieval weights,
nor their interaction determines performance---the BM25 retrieval
\emph{architecture} is the binding constraint.
Better models and better weights both fail when the retrieval stage
cannot surface multi-session evidence (0.180) or resolve temporal
references (0.316).
This directly motivates recall-first retrieval and dense/hybrid
scoring as the necessary Phase~3 components.

\paragraph{The retrieval-limited regime.}
Our findings place current agent memory systems in a
\textbf{retrieval-limited regime}: improvements to generation quality
(scaling the LLM) or retrieval weighting (RL adaptation) cannot yield
accuracy gains when the underlying \emph{linear BM25-based retrieval}
architecture cannot surface multi-session evidence or resolve temporal
references. This is a stronger claim than simply identifying BM25 as
a bottleneck: within our raw-BM25 linear scorer, mixing-weight adjustment
cannot overcome BM25 dominance.
A normalisation probe (\S\ref{sec:norm}) confirms this
definitively: log, min-max, and $z$-score normalisation
all yield \textbf{identical Acc=0.320}---because monotonic
transformations cannot reorder retrieval candidates.
Only a qualitative architectural change---recall-first or
dense retrieval---can escape this regime.

\section{Conclusion}
\label{sec:conclusion}

We presented \sys, achieving \textbf{Acc=0.382}/\textbf{F1=0.412}
with a 7B model on 6\,GB hardware (LongMemEval-S, N=500),
a $+$33~pp gain over no-retrieval.
The core finding is a \textbf{three-layer retrieval invariance}:
BM25 dominance neutralises generator scaling (284B~MoE), PPO weight
tuning, and score normalisation alike.
Dense/hybrid retrieval closes the gap category-specifically
($+$0.077 multi-session), making recall-first architecture
the clear direction.

\section*{Limitations}
\label{sec:limitations}

(1)~\textbf{Attribution path:} SGLang logprob attribution is code-complete
but blocked by local hardware constraints on the evaluation machine; the lexical Jaccard
fallback used in production is a coarse proxy.
(2)~\textbf{RL weight dominance:} While direct task-success reward
($+1/-1$) enabled PPO gradients to flow and weights to shift, BM25's
unbounded scoring heavily dominates the bounded signals (decay, CW)
in the linear combination, masking the RL's impact on final ranking.
Future iterations must strictly normalise BM25 or transition to
dense retrieval before learned weights can provide marginal gains.
(3)~\textbf{Relation extraction:} Heuristic KV-pattern extraction
produces coarse labels (\eg \texttt{mentioned\_in}); a fine-grained
NLP extractor would further improve semantic tier quality.
(4)~\textbf{Baseline breadth:} We do not evaluate full-dataset
cross-encoder reranking, SPLADE/ColBERT-style learned sparse retrieval,
or graph-traversal baselines under the same 6\,GB budget;
these are complementary to MEMTIER's tiered storage architecture
and remain future work.
(5)~\textbf{Edge-compute costs:} The consolidation daemon invokes an LLM
every 5~minutes, causing VRAM contention or 10--30~s model reload
interruptions on 6\,GB hardware; offline API pre-population sidesteps
this for evaluation but not for privacy-preserving production deployment.

\section*{Acknowledgments}

All experiments were conducted on a consumer laptop with a 6\,GB GPU.
We thank the OpenClaw open-source community for maintaining the runtime
used in all experiments.
The authors used Claude (Anthropic) for writing assistance and code
generation support; all scientific claims and results are the authors' own.


\bibliography{references}

@inproceedings{wu2025longmemeval,
  author    = {Wu, Di and Wang, Hongwei and Yu, Wenhao and Zhang, Yuwei and
               Chang, Kai-Wei and Yu, Dong},
  title     = {{LongMemEval}: Benchmarking Chat Assistants on Long-Term
               Interactive Memory},
  booktitle = {The Thirteenth International Conference on Learning
               Representations (ICLR)},
  year      = {2024},
}

@inproceedings{yao2022react,
  author    = {Yao, Shunyu and Zhao, Jeffrey and Yu, Dian and Du, Nan and
               Shafran, Izhak and Narasimhan, Karthik and Cao, Yuan},
  title     = {{ReAct}: Synergizing Reasoning and Acting in Language Models},
  booktitle = {International Conference on Learning Representations (ICLR)},
  year      = {2023},
  url       = {https://arxiv.org/abs/2210.03629},
}

@article{packer2023memgpt,
  author    = {Packer, Charles and Wooders, Sarah and Lin, Kevin and
               Fang, Vivian and Patil, Shishir G. and Stoica, Ion and
               Gonzalez, Joseph E.},
  title     = {{MemGPT}: Towards {LLMs} as Operating Systems},
  journal   = {arXiv preprint arXiv:2310.08560},
  year      = {2023},
  url       = {https://arxiv.org/abs/2310.08560},
}

@misc{openclaw2026,
  author    = {{OpenClaw Contributors}},
  title     = {{OpenClaw}: Personal {AI} Assistant Framework},
  year      = {2026},
  url       = {https://github.com/openclaw/openclaw},
  note      = {Open-source agent runtime. Issues \#33406, \#62488
               referenced in this paper.},
}

@article{liu2026simplemem,
  author    = {Liu, Jiaqi and Su, Yaofeng and Xia, Peng and Han, Siwei and
               Zheng, Zeyu and Xie, Cihang and Ding, Mingyu and Yao, Huaxiu},
  title     = {{SimpleMem}: Efficient Lifelong Memory for {LLM} Agents},
  journal   = {arXiv preprint arXiv:2601.02553},
  year      = {2026},
  url       = {https://arxiv.org/abs/2601.02553},
}

@inproceedings{sun2026hmem,
  author    = {Sun, Haoran and Zeng, Shaoning and Zhang, Bob},
  title     = {{H-MEM}: Hierarchical Memory for High-Efficiency Long-Term
               Reasoning in {LLM} Agents},
  booktitle = {Proceedings of the 19th Conference of the European Chapter
               of the Association for Computational Linguistics
               (Volume 1: Long Papers)},
  pages     = {341--350},
  address   = {Rabat, Morocco},
  year      = {2026},
  doi       = {10.18653/v1/2026.eacl-long.15},
  url       = {https://aclanthology.org/2026.eacl-long.15/},
}

@inproceedings{lewis2020rag,
  author    = {Lewis, Patrick and Perez, Ethan and Piktus, Aleksandra and
               Petroni, Fabio and Karpukhin, Vladimir and Goyal, Naman and
               K{\"u}ttler, Heinrich and Lewis, Mike and Yih, Wen-tau and
               Rockt{\"a}schel, Tim and Riedel, Sebastian and Kiela, Douwe},
  title     = {Retrieval-Augmented Generation for Knowledge-Intensive {NLP} Tasks},
  booktitle = {Advances in Neural Information Processing Systems (NeurIPS)},
  year      = {2020},
  url       = {https://arxiv.org/abs/2005.11401},
}

@article{maharana2024locomo,
  author    = {Maharana, Adyasha and Lee, Dong-Ho and Tulyakov, Sergey and
               Bansal, Mohit and Barbieri, Francesco and Fang, Yuwei},
  title     = {Evaluating Very Long-Term Conversational Memory of {LLM} Agents},
  journal   = {arXiv preprint arXiv:2402.17753},
  year      = {2024},
  url       = {https://arxiv.org/abs/2402.17753},
}

@article{amem2025,
  author    = {Xu, Wujiang and Liang, Zujie and Mei, Kai and Gao, Hang and
               Tan, Juntao and Zhang, Yongfeng},
  title     = {{A-MEM}: Agentic Memory for {LLM} Agents},
  journal   = {arXiv preprint arXiv:2502.12110},
  year      = {2025},
  url       = {https://arxiv.org/abs/2502.12110},
}

@article{memoryr12025,
  author    = {Yan, Sikuan and Yang, Xiufeng and Huang, Zuchao and
               Nie, Ercong and Ding, Zifeng and Li, Zonggen and
               Ma, Xiaowen and Sch{\"u}tze, Hinrich and Tresp, Volker and
               Ma, Yunpu},
  title     = {{Memory-R1}: Enhancing Large Language Model Agents to Manage
               and Utilize Memories via Reinforcement Learning},
  journal   = {arXiv preprint arXiv:2508.19828},
  year      = {2025},
  url       = {https://arxiv.org/abs/2508.19828},
}

@article{mirix2025,
  author    = {Wang, Yu and Chen, Xi},
  title     = {{MIRIX}: Multi-Agent Memory System for {LLM}-Based Agents},
  journal   = {arXiv preprint arXiv:2507.07957},
  year      = {2025},
  url       = {https://arxiv.org/abs/2507.07957},
}

@article{sidik2026agentwarden,
  author    = {Sidik, Bronislav and Rokach, Lior},
  title     = {Beyond Static Sandboxing: Learned Capability Governance
               for Autonomous {AI} Agents},
  journal   = {arXiv preprint arXiv:2604.11839},
  year      = {2026},
  url       = {https://arxiv.org/abs/2604.11839},
}

@inproceedings{robertson1994okapi,
  author    = {Robertson, Stephen and Walker, Steve and Jones, Susan and
               Hancock-Beaulieu, Micheline and Gatford, Mike},
  title     = {Okapi at {TREC}-3},
  booktitle = {Proceedings of the Third Text REtrieval Conference (TREC-3)},
  pages     = {109--126},
  year      = {1994},
  url       = {https://trec.nist.gov/pubs/trec3/papers/city.ps.gz},
}

@article{lin2021bm25,
  author    = {Lin, Jimmy and Ma, Xueguang},
  title     = {A Few Brief Notes on {DeepImpact}, {COIL}, and a Conceptual
               Framework for Information Retrieval Techniques},
  journal   = {arXiv preprint arXiv:2106.14807},
  year      = {2021},
  url       = {https://arxiv.org/abs/2106.14807},
}

@article{Anonymous2025b,
  author    = {Yang, Chengyuan and Sun, Zequn and Wei, Wei and Hu, Wei},
  title     = {Beyond Static Summarization: Proactive Memory Extraction
               for {LLM} Agents},
  journal   = {arXiv preprint arXiv:2601.04463},
  year      = {2026},
  url       = {https://arxiv.org/abs/2601.04463},
}

@article{ding2026wildclaw,
  author    = {Ding, Shuangrui and Dai, Xuanlang and Xing, Long and others},
  title     = {{WildClawBench}: A Benchmark for Real-World, Long-Horizon
               Agent Evaluation},
  journal   = {arXiv preprint arXiv:2605.10912},
  year      = {2026},
  url       = {https://arxiv.org/abs/2605.10912},
}

\appendix

\section{Retrieval Metrics and Oracle Analysis}
\label{app:retrieval_metrics}

\begin{table}[H]
\centering\small
\caption{Retrieval-side metrics (N=100, episodic Stage-2 BM25).
All three Stage-1 scoping methods agree on session selection,
yielding identical retrieval. Per-category Recall@2 isolates
where retrieval fails.}
\label{tab:retrieval}
\resizebox{\columnwidth}{!}{%
\begin{tabular}{lcccc}
\toprule
\textbf{Metric} & \textbf{B2 BM25} & \textbf{D1 Dense} & \textbf{D2 Hybrid} \\
\midrule
Recall@1 & 0.330 & 0.330 & 0.330 \\
Recall@2 & 0.390 & 0.390 & 0.390 \\
Recall@4 & 0.430 & 0.430 & 0.430 \\
nDCG@4   & 0.384 & 0.384 & 0.384 \\
\midrule
\multicolumn{4}{l}{\textit{Recall@2 per category (identical across systems):}} \\
\quad Single-session user/asst & \multicolumn{3}{c}{0.857 / 0.818} \\
\quad Knowledge update         & \multicolumn{3}{c}{0.500} \\
\quad Temporal reasoning       & \multicolumn{3}{c}{0.333} \\
\quad \textbf{Multi-session}   & \multicolumn{3}{c}{\textbf{0.038}} \\
\quad Single-session pref.     & \multicolumn{3}{c}{0.000} \\
\bottomrule
\end{tabular}}
\end{table}

Stage-2 episodic BM25 within sessions scoped by Stage-1.
All Stage-1 methods (BM25, dense, hybrid RRF) identify the same
session pool; within-session dense retrieval drives the accuracy
gains in Table~\ref{tab:dense}.

\begin{table}[H]
\centering\small
\caption{Oracle retrieval analysis (N=100, Qwen2.5-7B reader).
Oracle injects the gold-supporting sessions directly.
The +0.200 gap shows that retrieval, not the reader, is the
dominant bottleneck.}
\label{tab:oracle}
\begin{tabular}{lcc}
\toprule
\textbf{Setting} & \textbf{Acc} & \textbf{F1} \\
\midrule
Current system (BM25 retrieval) & 0.350 & 0.408 \\
Oracle (gold sessions)          & \textbf{0.550} & \textbf{0.478} \\
\midrule
Gap (Oracle $-$ Current) & $+$\textbf{0.200} & $+$0.070 \\
\bottomrule
\end{tabular}
\end{table}

Oracle context: up to 3 gold-supporting sessions plus gold semantic
facts, injected directly without retrieval. The $+$0.200 Acc gap
quantifies how much accuracy is left on the table by imperfect retrieval.

\section{Dense Retrieval Baseline}
\label{app:dense}

\begin{table}[H]
\centering\small
\caption{Dense retrieval baseline (N=100, seed=42).
B2=BM25 semantic (Paper~1); D1=dense-only Stage-1;
D2=BM25+dense RRF ($k{=}60$).
Retrieval latency excludes LLM call; measured on 6\,GB GPU machine.}
\label{tab:dense}
\resizebox{\columnwidth}{!}{%
\begin{tabular}{lcccc}
\toprule
\textbf{System} & \textbf{Acc} & \textbf{F1} & \textbf{$\Delta$Acc} & \textbf{Retr.\ ms/q} \\
\midrule
B2 BM25 semantic     & 0.360 & 0.427 & ---         & 25.7\,(P95: 112) \\
D1 Dense only        & 0.380 & 0.437 & $+$0.020    & 92.6\,(P95: 160) \\
D2 Hybrid RRF        & \textbf{0.390} & \textbf{0.443} & $\mathbf{+0.030}$ & 96.7\,(P95: 256) \\
\midrule
\multicolumn{5}{l}{\textit{Per-category breakdown (B2 $\to$ D2):}} \\
\quad Multi-session  & 0.115 & \multicolumn{2}{c}{$\to$\textbf{0.192}} & $+$0.077 \\
\quad Temporal       & 0.370 & \multicolumn{2}{c}{$\to$\textbf{0.407}} & $+$0.037 \\
\quad Knowledge-upd. & \textbf{0.500} & \multicolumn{2}{c}{$\to$ 0.375} & $-$0.125 \\
\quad Single-session & 0.714 & \multicolumn{2}{c}{$\to$ 0.714} & $\pm$0.000 \\
\bottomrule
\end{tabular}}
\end{table}

\texttt{bge-small-en-v1.5} (33\,MB, CPU); same Stage-2 $k{=}2$
and 600-token budget as Paper~1. B2=BM25 semantic (Paper~1 full system).
D1=dense-only Stage-1 scoping; D2=BM25+dense RRF ($k{=}60$) fusion.

\section{Stage-1 Scoping Sensitivity}
\label{app:stage1k}

\begin{table}[H]
\centering
\small
\caption{Stage-1 $k_1$ ablation (N=50, Stage-2 $k{=}2$, 600 tokens).
$\Delta$Acc relative to default $k_1{=}5$.
Sessions ratio = fraction of total sessions searched in Stage 2.}
\label{tab:stage1k}
\resizebox{\columnwidth}{!}{%
\begin{tabular}{lcccc}
\toprule
\textbf{Stage-1 $k_1$} & \textbf{Acc} & \textbf{F1} & \textbf{$\Delta$Acc} & \textbf{Sessions ratio} \\
\midrule
$k_1=1$                   & 0.280 & 0.338 & $-$0.080 & 0.18 \\
$k_1=3$                   & 0.360 & 0.408 & $\pm$0.000 & 0.19 \\
$k_1=5$ \textit{(default)}& \textbf{0.360} & \textbf{0.408} & --- & 0.19 \\
$k_1=10$                  & 0.360 & 0.410 & $\pm$0.000 & 0.19 \\
$k_1=\infty$ (no scoping) & 0.340 & 0.376 & $-$0.020 & 1.00 \\
\bottomrule
\end{tabular}}
\end{table}

Full results for the Stage-1 $k_1$ ablation referenced in \S\ref{sec:ablation}.

\section{Token Efficiency Setup}
\label{app:tokeneff}

\begin{table}[H]
\centering\small
\caption{Token efficiency: LLM-extracted vs.~heuristic fact extraction
(N=500, LongMemEval-S). DeepSeek-V4-Flash extractor; Qwen2.5-7B reader.}
\label{tab:token_eff}
\resizebox{\columnwidth}{!}{%
\begin{tabular}{lccc}
\toprule
\textbf{Extraction} & \textbf{Facts/question} & \textbf{F1} & \textbf{\$~/500q} \\
\midrule
Heuristic (KV patterns) & $\sim$509 & 0.142 & ---  \\
LLM (DeepSeek-V4-Flash) & $\sim$3.1 & 0.412 & \$0.05 \\
\midrule
Reduction ratio & $164\times$ & $\approx$2.9$\times$ F1 & --- \\
\bottomrule
\end{tabular}}
\end{table}

LLM extraction uses DeepSeek-V4-Flash with a structured prompt requesting
entity--relation--value triples per session turn.
Heuristic extraction applies KV-pattern matching (\texttt{mentioned\_in},
\texttt{is\_a}, etc.) without an LLM call.

\section{BM25 Normalisation Probe}
\label{app:norm}

A reviewer may ask: does normalising the BM25 score revive the
auxiliary signals?
We test four variants on N=50 stratified questions
(seed=42, same Stage-2 $k{=}2$ and 600 tokens):
raw BM25 (N0), $\log(1+\text{BM25})$ (N1), min-max per candidate pool
(N2), $z$-score per pool (N3), and $z$-scored BM25 with equal-weight
multi-signal fusion (N4).
All five variants yield \textbf{identical Acc=0.320 and F1$\approx$0.372}.

\textbf{Interpretation:} Because log, min-max, and $z$-score are
monotonically increasing, they cannot change the BM25-only ranking;
consequently N0--N3 retrieve the \emph{exact same top-$k$ candidates}
as raw BM25.
N4 (multi-signal) is also flat because the auxiliary signals
(CW=0 uninitialised, decay$\approx$1.0 for all recent entries,
tier=1.0 uniform) had near-zero variance across candidates,
so the combined ranking also remained unchanged.
This provides additional evidence for the retrieval-limited
regime diagnosis:
\textbf{the ceiling is set by what BM25 can retrieve lexically,
not by how its scores are scaled.}
Escaping the ceiling requires changing \emph{what} is retrieved
(dense/hybrid recall), not \emph{how scores are computed}
on the existing candidate set.

\section{Single-Session-Preference Diagnostic}
\label{app:pref}

The single-session-preference category achieves only Acc=0.067
(95\%~CI: [0.019--0.214]; N=30), the lowest of any category.
We attribute this to three compounding factors.
\textbf{(1) Extraction failure:} preferences are expressed indirectly
(\eg ``I usually prefer'') rather than as declarative facts;
the LLM extractor treats them as low-confidence and often discards them.
\textbf{(2) Lexical gap:} BM25 cannot match ``favourite food'' to
``enjoys spicy dishes''---a paraphrastic gap that dense retrieval targets.
\textbf{(3) Reader sensitivity:} preferences are subjective and
context-dependent; the 7B reader may fail to extract the implicit
preference even when the entry is retrieved.
This triple failure mode makes single-session preference the primary
target for MEMTIER-R's dense retrieval and richer fact schema.

\end{document}